\documentclass[twoside]{article}

\usepackage[preprint]{aistats2026}
\usepackage{graphicx}
\usepackage{booktabs}
\usepackage{amssymb}
\usepackage{framed}
\usepackage{dsfont}
\usepackage{url}

\usepackage[round]{natbib}

\begin{document}

\runningtitle{From Fine-Grained Entropy to Calibrated Uncertainty}
\runningauthor{Jenane, Walha, Kuhn, Buettner}

%

%

\twocolumn[

\aistatstitle{From Entropy to Calibrated Uncertainty:\\ Training Language Models to Reason About Uncertainty}

\aistatsauthor{
Azza Jenane$^{*}$ ,
Nassim Walha$^{*}$,
Lukas Kuhn, 
Florian Buettner
}

\aistatsaddress{
German Cancer Research Center (DKFZ) \\
German Cancer Consortium (DKTK) \\
Goethe University Frankfurt, Germany
}

]
\begin{abstract}
Large Language Models (LLMs) that can express interpretable and calibrated uncertainty are crucial in high-stakes domains. While methods to compute uncertainty post-hoc exist, they are often sampling-based and therefore computationally expensive or lack calibration. We propose a three-stage pipeline to post-train LLMs to efficiently infer calibrated uncertainty estimates for their responses. First, we compute fine-grained entropy-based uncertainty scores on the training data, capturing the distributional variability of model outputs in embedding space. Second, these scores are calibrated via Platt scaling, producing reliable and human-interpretable uncertainty signals. Finally, the target LLM is post-trained via reinforcement learning to align its policy with these calibrated signals through a verifiable reward function. Unlike post-hoc uncertainty estimation methods, our approach provides interpretable and computationally efficient uncertainty estimates at test time. Experiments show that models trained with our pipeline achieve better calibration than baselines and generalize to unseen tasks without further processing, suggesting that they learn a robust uncertainty reasoning behavior. 
\end{abstract}

\footnotetext[1]{* Equal contribution.}

\begin{figure}[h!]
\vspace{.3in}
\centerline{\includegraphics[width=0.9\linewidth]{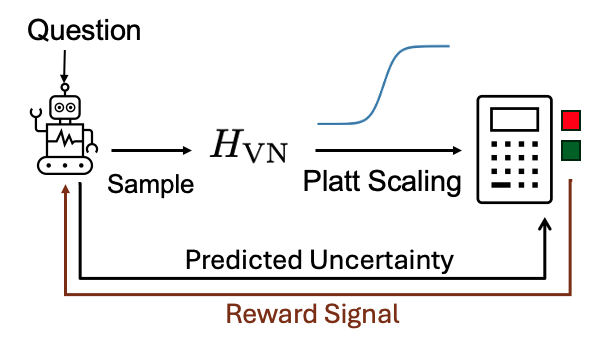}}
\vspace{.3in}
\caption{Overview of the reward signal generation pipeline}
\label{fig:data_gen}
\end{figure}

\section{Introduction}

Large Language Models (LLMs) have achieved strong performance across a broad range of Natural Language Processing (NLP) tasks, including question answering, summarization, and natural language understanding \citep{touvron2023llama, chiang2023vicuna, achiam2023gpt}. Despite these advances, they remain prone to generating confident yet incorrect outputs, commonly referred to as hallucinations \citep{hadi2023survey}. 

As LLMs are increasingly deployed in high-stakes domains such as healthcare \citep{wang2024large, bani2024magda}, finance \citep{cao2022ai}, and legal decision support \citep{zhong2020does}, reliable and interpretable uncertainty estimates are essential to enable risk-aware decision-making, appropriate human oversight, and safe system integration. A central requirement for such reliability is to have calibrated confidence; predicted uncertainty must align with empirical correctness \citep{guo2017calibration}.

Current approaches to uncertainty estimation in LLMs are predominantly post-hoc, relying on sampling-based statistics and entropy measures \citep{farquhar2024detecting, walha2025fine, nikitin2024kernel}. Given a user query, these methods generate multiple responses from the model and quantify uncertainty through the semantic variability among the sampled outputs. While effective, they incur substantial computational overhead due to repeated sampling. Moreover, they yield scale-free uncertainty values that perform well on ranking-based metrics such as AUROC and AUPR, but are inherently uncalibrated, as they do not map directly to probabilities. A separate line of work leverages verbalized uncertainty, employing various prompting strategies to elicit explicit confidence scores from the model \citep{kadavath2022language, xiong2023can}. These methods are generally more computationally efficient, but their reliability is strongly dependent on model size and capacity. In particular, smaller models deployed on-device in privacy-sensitive settings have been shown to produce poorly calibrated and unreliable confidence estimates \citep{xiong2023can, yang2024verbalized}. More recent reinforcement learning approaches introduce verifiable rewards to encourage alignment between predicted confidence and actual correctness; however, they often rely on coarse supervision signals or computationally expensive optimization schemes \citep{bani2025rewarding, damani2025beyond}.

To address the aforementioned limitations, we propose a three-stage framework that post-trains LLMs to express calibrated uncertainty directly. In the first stage, we compute fine-grained uncertainty scores using von Neumann entropy \citep{von2018mathematical, walha2025fine} over embedding representations derived from multiple sampled generations, capturing distributional structure beyond binary correctness. In the second stage, these scores are calibrated via Platt scaling \citep{platt1999probabilistic} to obtain interpretable probability targets. In the third stage, the model is trained using Group Relative Policy Optimization (GRPO) with a verifiable reward \citep{wen2025reinforcement, shao2024deepseekmath} that aligns the model's predicted uncertainty with these calibrated signals. This framework integrates uncertainty estimation directly into model behavior while remaining computationally efficient at inference time. The contributions of this work are as follows:

\begin{itemize}
    \item We introduce a novel uncertainty calibration reward that aligns the model's verbalized uncertainty with a state-of-the-art sampling-based measure, while explicitly targeting calibrated probability outputs.
    \item We demonstrate that our reward yields verbalized uncertainties with high rank-correlation with the sampling-based measure, thus inheriting its strong performance on ranking-based metrics, while also achieving state-of-the-art calibration and high efficiency at inference time.
    \item We compare our reward against a Brier-score-based reward \citep{brier1950verification} commonly used in the literature \citep{damani2025beyond}, and demonstrate superior performance both in-distribution and out-of-distribution. 
\end{itemize}

\section{Background}
\subsection{Fine-Grained Entropy-Based Uncertainty}
\citet{walha2025fine} propose a fine-grained uncertainty measure based on spectral entropy in embedding space. For each input, multiple generations are sampled and mapped to embedding vectors. A kernel matrix is constructed over the embeddings to capture pairwise similarity between generated responses.

The eigenvalues $\{\lambda_i\}_{i=1}^N$ of the normalized kernel matrix are then used to compute the von Neumann entropy \citep{von2018mathematical},
\begin{equation*}
H_{\mathrm{VN}} = - \sum_{i=1}^{N} \lambda_i \log \lambda_i,
\end{equation*}
which quantifies dispersion in representation space.  By operating at the representation level and aggregating across multiple samples, this approach provides a continuous uncertainty signal that captures distributional variability in semantic space, beyond token-level predictive entropy or binary correctness signals.

\subsection{Platt Scaling for Calibration}
\paragraph{Calibrated confidence \citep{guo2017calibration}.}
Let $\hat{Y}$ and $\hat{P}$ denote the model's predicted answer and its associated confidence, respectively, and let $Y$ denote the ground truth answer to query $X$. A calibrated model satisfies
\begin{equation*}
    \mathbb{P}(\hat{Y}=Y \mid \hat{P}=p) \approx p, \quad \forall p\in[0,1],
\end{equation*}
meaning that among all predictions for which the model assigns confidence $p$, the fraction of correct predictions is approximately $p$.

\paragraph{Platt scaling \citep{platt1999probabilistic}.}
Platt scaling is a post-hoc calibration method that maps arbitrary real-valued scores to calibrated probability estimates via a parametric logistic transformation. Given an uncalibrated score $s \in \mathbb{R}$, Platt scaling fits a sigmoid function
$
p = \sigma(As + B),
$
where the parameters $A$ and $B$ are learned by minimizing the negative log-likelihood on a held-out validation set with binary correctness labels. This transformation preserves the ranking of the original scores while rescaling their values so that the resulting probabilities better reflect empirical accuracy.

\subsection{Reinforcement Learning for Uncertainty Estimation}

Reinforcement Learning (RL) provides a framework for optimizing model behavior with respect to a reward signal rather than supervised targets. It has recently been explored for uncertainty estimation by defining rewards that encourage alignment between predicted confidence and empirical correctness \citep{damani2025beyond, bani2025rewarding}. Such formulations allow uncertainty calibration to be integrated directly into the training objective, rather than computed post-hoc.


Group Relative Policy Optimization (GRPO) has been proposed as a computationally efficient reinforcement learning algorithm particularly suited for training LLMs \citep{shao2024deepseekmath}. Unlike Proximal Policy Optimization (PPO), which requires maintaining and updating a large critic network throughout training, GRPO performs relative comparisons within groups of sampled responses, using group-normalized rewards to guide policy updates. This substantially reduces both memory usage and training complexity while preserving training stability. As a result, GRPO enables scalable reinforcement learning for LLMs at a lower computational cost compared to standard PPO-based pipelines.
 \section{Methodology}
Our pipeline, illustrated in Figure \ref{fig:data_gen}, aims to improve the LLM's ability to accurately estimate its own uncertainty given an input question and the model's corresponding answer. Formally, let $\mathcal{X}$ denote the space of questions and $\mathcal{Y}$ denote the space of answers. Given a question $x \in \mathcal{X}$ with ground truth answer $y^* \in \mathcal{Y}$, we sample an answer $\hat{y}$ from the base LLM $\pi_{\theta_0}(y \mid x)$. We then fine-tune the model to predict a scalar uncertainty estimate $u_\theta(x, \hat{y}) \in [0,1]$, interpreted as the probability that the answer $\hat{y}$ is incorrect, i.e., $u_\theta(x, \hat{y}) \approx \mathbb{P}(\hat{y} \neq y^\ast \mid x)$.

\paragraph{Entropy-Based Uncertainty Signal.}

Following the work of \cite{walha2025fine}, for a fixed base model $\pi_{\theta_0}$, we generate $K$ stochastic samples $\{ y^{(k)} \}_{k=1}^K \sim \pi_{\theta_0}(\cdot \mid x)$ and compute a semantic dispersion score 
\begin{equation*}
    S(x) = H_{VN}(x)
\end{equation*}

using kernel-based von Neumann entropy over their embedding representations. 
This score serves as a continuous proxy for uncertainty.

\paragraph{Calibration Mapping.}
Since $S(x)$ is not inherently probabilistic and raw values are not directly interpretable or calibrated, we learn a calibration function 
$g : \mathbb{R}_{\geq 0} \to [0,1]$ using Platt scaling \citep{platt1999probabilistic} on held-out data with binary correctness labels $z \in \{0,1\}$, where $z=1$ indicates an incorrect response. The calibrated uncertainty is then defined as $u_{\text{cal}}(x) = g(S(x))$, which estimates $\mathbb{P}(\text{incorrect} \mid x)$ under the validation distribution. 

\paragraph{Reinforcement Learning for Calibrated Uncertainty.}

Similar to \citet{bani2025rewarding}, we decouple answer generation from uncertainty estimation during training: answers are generated first and treated as fixed, while uncertainty is produced in a separate step and optimized independently. This ensures that answer quality remains unaffected by the uncertainty calibration objective. Unlike \citet{damani2025beyond}, which fine-tunes all model weights, we adopt Low-Rank Adaptation (LoRA) \citep{hu2022lora} as a parameter-efficient alternative. Beyond reducing memory overhead and mitigating catastrophic forgetting, LoRA naturally supports the decoupling of answer generation and uncertainty prediction at inference time, as the learned adapters can be selectively applied after answer generation to produce uncertainty estimates.

To further optimize training efficiency, we use reinforcement learning with the Group Relative Policy Optimization (GRPO) algorithm. Our entropy-based reward function encourages alignment between the predicted uncertainty $u_\theta(x, \hat{y})$ and the calibrated target $u_{\text{cal}}(x)$ and is defined as follows:
\[
R_{\textbf{entropy}}(u_\theta, u_{\text{cal}}) = 1 - \max\big(0.05 , \; |u_\theta - u_{\text{cal}}|\big).
\]
During training, the model is provided with the question $x$ and its pre-generated answer $\hat{y}$, and is prompted to first produce a reasoning trace about its uncertainty in a chain-of-thought (CoT) format, followed by a scalar uncertainty prediction $u_\theta(x, \hat{y})$. By optimizing the scalar output directly, the model is implicitly encouraged to develop a useful reasoning trace that supports reliable uncertainty estimation.

\section{Experiments}

\subsection{Experimental Setup}
\begin{table*}[t]
\centering
\caption{Performance metrics across methods on in-domain (TriviaQA + Natural Questions) and out-of-domain (GSM8k) datasets.}
\label{tab:metrics}
\resizebox{0.95\textwidth}{!}{%
\begin{tabular}{lccc|cc}
\toprule
& \multicolumn{3}{c}{\textbf{TriviaQA + NQ (ID)}} 
& \multicolumn{2}{c}{\textbf{GSM8k (OOD)}} \\
\cmidrule(lr){2-4} \cmidrule(lr){5-6}
\textbf{Method} 
& \textbf{ECE (\%) $\downarrow$} 
& \textbf{AUROC (\%) $\uparrow$} 
& \textbf{Spearman $\uparrow$}
& \textbf{ECE (\%) $\downarrow$} 
& \textbf{AUROC (\%) $\uparrow$} \\
\midrule
\textbf{Base}  & 41.99 & 51.89 & 0.03 & 32.22 & 53.79 \\
\textbf{Base+CoT} & 34.17 & 66.18 & 0.17 & 22.25 & 62.17 \\
\textbf{Brier} & 15.70 & \textbf{83.36} & 0.52 & 33.28 & \textbf{66.89} \\
\textbf{Entropy-based (ours)} & \textbf{7.2} & 81.53 & \textbf{0.67} & \textbf{3.15} & \textbf{66.73} \\
\bottomrule
\end{tabular}%
}
\end{table*}
We conduct experiments on subsets of TriviaQA \citep{joshi2017triviaqa} and Natural Questions (NQ) \citep{kwiatkowski2019natural}, two widely used open-domain question answering benchmarks. We adopt the Qwen2.5-7B-Instruct model, which has demonstrated strong performance in reinforcement learning settings \citep{hu2025open, damani2025beyond}, and initialize reinforcement learning training directly from this base model, following recent practice. To assess the performance of our method and the baselines, we report Expected Calibration Error (ECE), Area Under the Receiver Operating Characteristic Curve (AUROC), and Spearman correlation with respect to the calibrated uncertainty targets. Full experimental details, including evaluation protocol, training procedure, and hyperparameter settings, are provided in Appendix \ref{sec:appendix_impl}.

\paragraph{Methods.}
We evaluate the following methods, which differ in their reward design:

\begin{itemize}
    \item \textbf{Base}: The pretrained Qwen2.5-7B-Instruct model without reinforcement learning, serving as a reference.

     \item \textbf{Base + CoT}: The pretrained Qwen2.5-7B-Instruct model without reinforcement learning, aided by chain-of-thought reasoning \citep{wei2022chain}.
     
    \item \textbf{Brier}: Initialized from the base model and trained with GRPO using a reward derived solely from the Brier score between predicted uncertainty and binary correctness labels. This reward is defined as $R_{{Brier}}(u_{\theta})=1-(u_{\theta}-\mathds{1}_{\hat{y}\neq y^*})^2$ and was used in previous work \citep{damani2025beyond}.
    
    \item \textbf{Entropy-based (ours)} : Initialized from the base model and trained with GRPO using our calibrated entropy-based reward signal $R_{entropy}$.
    
\end{itemize}

\subsection{Results}

We evaluate all methods on a held-out in-domain (ID) test set comprising TriviaQA and Natural Questions, and assess out-of-domain (OOD) generalization on GSM8K \citep{cobbe2021gsm8k}. Results are reported in Table~\ref{tab:metrics}.

\paragraph{In-domain performance.}
Our entropy-based RL method achieves the strongest overall performance. It reduces ECE from 41.99\% (Base) and 34.17\% (Base+CoT) to \textbf{7.2\%}, substantially improving calibration over all baselines. While the Brier variant also improves calibration, reaching 15.70\%, it remains notably worse than the entropy-based approach. In terms of ranking quality, both the Brier (83.36\%) and entropy-based (81.53\%) variants substantially outperform the base models (51.89\% and 66.18\%, respectively). Furthermore, the entropy-based method achieves the highest alignment with calibrated uncertainty signals, yielding the strongest Spearman correlation (\textbf{0.67}) across all methods.

\paragraph{Out-of-domain generalization.}
On GSM8K, the entropy-based method again achieves the best calibration, reducing ECE to \textbf{3.15\%}, compared to 32.22\% (Base), 22.25\% (Base+CoT), and 33.28\% (Brier). In terms of AUROC, both the Brier (66.89\%) and entropy-based (66.73\%) variants substantially outperform the baselines (53.79\% and 62.17\%), with the Brier variant marginally higher.

Overall, while CoT prompting and Brier-based optimization improve ranking performance, they do not consistently yield well-calibrated uncertainty estimates. In contrast, reinforcement learning with entropy-calibrated uncertainty targets leads to substantial gains in both calibration and ranking quality, and generalizes robustly to out-of-domain settings.

\section{Discussion and Conclusion}
We introduced a novel uncertainty calibration reward grounded in an established entropy-based uncertainty measure for LLMs. Our approach not only produces uncertainty estimates that are well-aligned with this state-of-the-art measure, but also significantly outperforms several baselines in terms of both calibration and uncertainty ranking, with consistent gains observed in-distribution and out-of-distribution. Furthermore, by combining GRPO with LoRA adapters, our framework maximizes training efficiency while remaining computationally lightweight at inference time.

While we provide evaluation across multiple metrics and datasets, extending the experiments to a broader set of models would yield a more comprehensive assessment. Additionally, our evaluation remains purely empirical, which is common practice in LLM research but leaves open the question of theoretical grounding.

Overall, our approach represents a promising direction toward efficient and reliable uncertainty quantification for large language models.

\newpage
\bibliography{references}

\clearpage
\onecolumn
\appendix
\appendix

\section{Implementation Details}
\label{sec:appendix_impl}

\subsection{Calibrated uncertainty signals}
To compute entropy-based uncertainty signals, we sample model responses at temperature $t=1.0$, which balances output diversity and determinism for reliable variability estimation. For the correctness labels used in Platt scaling and evaluation, we follow \citet{walha2025fine} and \citet{farquhar2024detecting} and generate answers at a low temperature $t=0.1$, approximating the model's best-effort prediction. These answers are then compared against the ground truth using GPT-4o-mini \citep{openai2024gpt4omini} as an automated judge to obtain binary correctness labels.

\subsection{GRPO training}
We train on a combined subset of TriviaQA and Natural Questions consisting of 18{,}000 training samples and evaluate on a held-out set of 2{,}000 samples. All experiments are conducted on a single NVIDIA H200 NVL GPU, with total training time ranging between 10 and 14 hours depending on the method.

Reinforcement learning is performed using GRPO for 1{,}000 update steps with a batch size of 32 and a group size of 16. The grouped sampling stabilizes learning under stochastic reward signals derived from calibrated uncertainty targets. Within each group, responses are sampled with temperature 1.5 to encourage diversity and improve exploration during policy optimization. 

We employ LoRA-based parameter-efficient fine-tuning with rank $r=16$ and scaling factor $\alpha=32$, together with a dropout rate of 0.05. This configuration provides sufficient adaptation capacity while preserving the pretrained model’s generation abilities and preventing catastrophic forgetting.

\subsection{Evaluation}
We evaluate uncertainty quality using the following metrics:

\begin{itemize}
    \item \textbf{Expected Calibration Error (ECE)} $\downarrow$: measures the discrepancy between predicted uncertainty and empirical error rates. 
    Lower values indicate better calibration.
    \item \textbf{AUROC} $\uparrow$: Area Under the Receiver Operating Characteristic curve, which reflects the quality of the uncertainty ranking.
    \item \textbf{Spearman Correlation} $\uparrow$: is a rank correlation measure that describes the monotonic relationship between predicted uncertainty and calibrated uncertainty targets.
\end{itemize}

For evaluation, we generate standard answers at low temperature $t=0.1$ to approximate each model's best-effort prediction. These answers are provided as input to the uncertainty inference prompt (Figures \ref{fig:training_prompt} and \ref{fig:inference_prompt}) for all evaluated methods, from which the models produce their predicted uncertainties $u_{\theta}$. Binary correctness labels are obtained by comparing the generated standard answers against the ground truth using GPT-4o-mini as a judge. These labels are then used alongside the predicted uncertainties to compute ECE and AUROC. Spearman correlation is computed by comparing the predicted uncertainties against the calibrated target uncertainties $u_{\mathrm{cal}}$ on the test set.

\newpage

\section{Prompts}
\label{sec:prompts}

\begin{figure*}[h]
\begin{framed}
\noindent \textbf{Objective}\newline
Answer the question below. Think briefly, then give your final answer.
Always provide an answer—if you are unsure or do not know the answer, give your best guess without indicating uncertainty. Your answer has to be short and precise.
\newline

\textbf{Output Format}\newline
R: $\langle \text{1-2 sentences of reasoning} \rangle $
\newline
A: $\langle \text{your final answer} \rangle $\newline
\newline \textbf{Task}
\newline
 \end{framed}
\caption{Answer generation prompt template for entropy sampling and standard answers}
  \label{fig:generation_prompt}

\end{figure*}

\begin{figure*}[h]
\begin{framed}
\noindent \textbf{Objective}\newline
In this task, you will receive a question. You will also receive a ground truth answer to the question and a model generated answer. Your goal is to compare the ground truth answer and the model generated answer in order to decide whether the model generated answer is correct or not.\newline

\textbf{Important Rules}\newline
1. The model generated answer is correct, when it is a valid answer to the question, and semantically equivalent to the ground truth answer. It does not necessarily need to overlap with the ground truth answer lexically.\newline
2. If the model generated answer contains more information (more specific) or less information (less specific) than the ground truth answer, but still correctly answers the question, then you should consider it correct.\newline
3. If you decide that the model generated answer is correct, say yes, otherwise say no.\newline
4. Your output should only contain your decision (yes or no). It should not contain any other text, explanation or reasoning.\newline
 \end{framed}
\caption{Correctness judge prompt template.}
  \label{fig:correctness_jude_prompt}

\end{figure*}

\begin{figure*}[h]
\begin{framed}
\noindent 
You are an assistant that estimates your total uncertainty about your answer to a question.\newline
\newline
I previously asked you to answer the question below, and you provided the answer below.
Now, I want you to reflect on your answer and estimate how uncertain you are about it.\newline
\newline
Uncertainty is the probability that your answer would be wrong.
\newline
Before giving the final value, briefly reflect on why your answer might be right and why it might be wrong.
Use your current judgment to produce a probability between 0 and 1.
\newline
\newline
\textbf{Output Format}\newline
Reflection: $\langle \text{1-2 sentences on why the answer might be right or wrong} \rangle $
\newline
Total\_uncertainty: $\langle \text{float between 0 and 1} \rangle$ \newline
\end{framed}
\caption{Training and inference prompt template for post-trained and CoT methods.}
  \label{fig:training_prompt}
\end{figure*}

\begin{figure*}[h]
\begin{framed}
\noindent 
Given the question and your answer below, estimate your uncertainty about the answer.
Only output the number, no explanations.
\newline
\textbf{Output Format}\newline
Total\_uncertainty: $\langle \text{float between 0 and 1} \rangle$ \newline
\end{framed}
\caption{Inference prompt template for the base model.}
  \label{fig:inference_prompt}
\end{figure*}

\end{document}